%% file: main.tex
\pdfoutput=1

\documentclass[11pt]{article}

\usepackage[final]{acl}

\usepackage{times}
\usepackage{latexsym}

\usepackage[T1]{fontenc}
\input{sec/00_definations}

\usepackage[utf8]{inputenc}

\usepackage{microtype}

\usepackage{inconsolata}

\title{\ourstitle: Fine-grained Evaluations of Hallucinations \\ in Large Vision-Language Models}

\author{Liqiang Jing$^1$, {\bf Ruosen Li}$^1$, {\bf Yunmo Chen}$^2$, {\bf Xinya Du}$^1$ \\   $^1$University of Texas at Dallas, $^2$Johns Hopkins University \\ jingliqiang6@gmail.com, xinya.du@utdallas.edu}

\begin{document}
\maketitle

\input{sec/0_abstract}    
\input{sec/1_intro}

\input{sec/2_related_work}

\input{sec/3_method}
\input{sec/4_meta}

\input{sec/5_atomic_eval}
\input{sec/6_conclusion}

\section*{Acknowledgements}
We thank the anonymous reviewers for valuable and insightful feedback.
This research is supported in part by the National Science Foundation CAREER Grant IIS-2340435 and an Amazon Research Award.  Any opinions, findings, and conclusions or recommendations expressed herein are those of the authors and do not necessarily represent the views, either expressed or implied, of the U.S. Government.

\section*{Limitations}

\ours\ focuses primarily on factual precision, ensuring that each piece of information in a generated text is supported by the visual input. However, it does not account for factual recall, meaning it doesn't penalize models for generating fewer facts. This can be seen as unfair, as there's often a trade-off between precision and recall. Additionally, the distinction between precision and recall can sometimes be unclear, as a generation may contain supported information but still miss important details. To address this, we suggest reporting \ours\ and the average length of generated text. We leave a more holistic approach for future work.

\bibliography{custom}



\input{sec/X_suppl}

\end{document}

%% file: sec/00_definations.tex
\definecolor{green}{rgb}{0.1,0.1,0.1}
\definecolor{chocolate}{HTML}{D2691E}
\definecolor{maroon}{HTML}{A00000}
\definecolor{indigo}{HTML}{4B0082}
\definecolor{green}{HTML}{008000}
\definecolor{cadmiumgreen}{rgb}{0.0, 0.42, 0.24}

\usepackage{multirow}
\usepackage{amssymb}
\usepackage{pifont}
\usepackage{wrapfig} 
\usepackage{picinpar}

\makeatletter
\newcommand*\myfontsize{%
  \@setfontsize\myfontsize{8}{9}%
}
\makeatother

\makeatletter
\newcommand*\mysmallfontsize{%
  \@setfontsize\mysmallfontsize{7.4}{8.4}%
}
\makeatother

\newcommand{\ours}{\textsc{FaithScore}}
\newcommand{\ourstitle}{\textsc{FaithScore}}

\newcommand{\ourslong}{\textbf{Faith}fulness to Atomic Image Facts \textbf{Score}}

\usepackage{adjustbox}

\newcommand{\myskip}[1]{}

\newcommand{\ie}{\emph{i.e., }}
\newcommand{\eg}{\emph{e.g., }}

\usepackage{booktabs}

%% file: sec/0_abstract.tex
\begin{abstract}
We introduce \ours\ (\ourslong), a reference-free and fine-grained evaluation metric that measures the faithfulness of the generated free-form answers from large vision-language models (LVLMs). The \ours\ evaluation first identifies sub-sentences containing descriptive statements that need to be verified, then extracts a comprehensive list of atomic facts from these sub-sentences, and finally conducts consistency verification between fine-grained atomic facts and the input image. Meta-evaluation demonstrates that our metric highly correlates with human judgments of faithfulness.  We collect two benchmark datasets  (i.e. LLaVA-1k and MSCOCO-Cap) for evaluating LVLMs instruction-following hallucinations.
We measure hallucinations in state-of-the-art LVLMs with \ours\ on the datasets. Results reveal that current systems are prone to generate hallucinated content unfaithful to the image, which leaves room for future improvements.  We hope our metric \ours\ can help evaluate future LVLMs in terms of faithfulness and provide insightful advice for enhancing LVLMs' faithfulness. 
\end{abstract}

%% file: sec/1_intro.tex
\begin{figure}[!ht]
    \centering
    \includegraphics[width=0.9\linewidth]{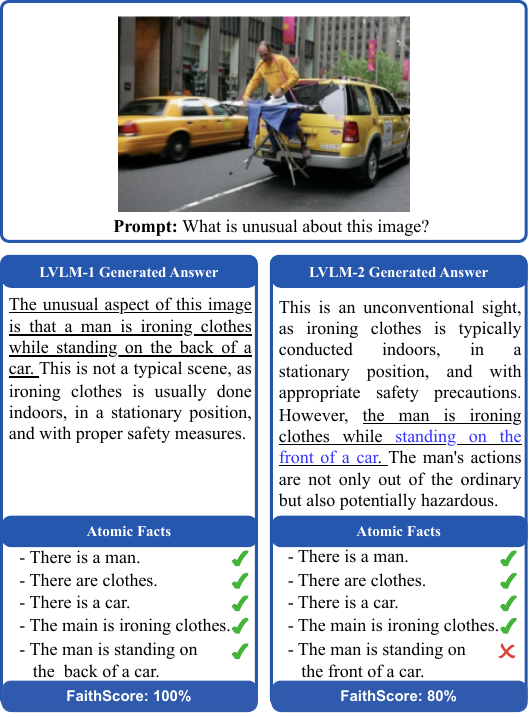}
    \caption{Illustration of how \ours\ evaluation works. Given the answers generated by an LVLM,
in step 1, we identify the descriptive content (with an LLM); 
In step 2, we extract corresponding atomic facts from the identified sentences; 
In step 3, the faithfulness of all atomic facts is verified according to the input image. 
In this example,  
the \underline{underlined part} denotes objective descriptive content in the answer.
The \textcolor{blue}{blue contents} denote hallucinations in the answers.
\ours\ allows a more
fine-grained and interpretable evaluation of the factual precision of free-form answers.
}
\vspace{-5mm}
    \label{fig:hallucination}
\end{figure}



\section{Introduction}


Large Language Models (LLMs), such as GPT-3~\cite{gpt3} and ChatGPT~\cite{chatgpt}, have demonstrated various language modeling capabilities. Despite their achievements, they still lack the capacity to handle multimodal inputs effectively. As a result, a significant amount of research has shifted its focus towards Large Vision-Language Models (LVLMs)~\cite{llava,mPLUG-Owl,llava_rlhf}
by incorporating powerful LLMs~\cite{llama,vicuna} and Vision Foundation Models~(VFMs)~\cite{vit,DBLP:journals/corr/abs-2108-07258}. 
LVLMs have shown strong performance on various multimodal tasks, such as Visual Question Answering~\cite{vqatask}, Image Captioning~\cite{mscoco}, and Multimodal Conversation~\cite{llava}.

Despite the effectiveness of LVLMs, the problem of hallucination is pervasive, often leading these models to generate fabricated information that is incongruent with the provided visual input~\cite{DBLP:conf/emnlp/RohrbachHBDS18,liu2023aligning,liu2023hallusionbench,DBLP:journals/corr/abs-2310-16045,chang2024unifiedhallucinationmitigationframework}. 
In the context of LVLM, the problem of hallucination can manifest as answers containing descriptions of the input image that are incorrect~\cite{Li2023EvaluatingOH}. 
As shown in Figure \ref{fig:hallucination}, the LVLM-2 generates an answer with an inaccurate description~(\ie \textit{standing on the front of a car}), which is not faithful towards the input image. The phenomenon of hallucination in LVLMs introduces potential hazards that could result in significant consequences such as misinformation and safety concerns, thus degrading the model's reliability in practical applications inevitably~\cite{DBLP:conf/chi/MacLeodBMC17}. Hence, it is imperative that these issues are thoroughly measured and addressed~\cite{ji2023survey}. 
%


Nevertheless, there have been limited explorations that 
measure the hallucination problem in LVLMs. \citet{Li2023EvaluatingOH} was among the first to measure the hallucinations of LVLMs with a polling-based object probing evaluation method. 
In addition, \citet{gunjal2023detecting} annotated a multi-modal hallucination detection dataset tailored for detailed image description evaluation. \citet{lovenia2023negative} devised Negative Object Presence Evaluation (NOPE), which used VQA to quantitatively evaluate object hallucination in LVLMs. 
These approaches, however, exhibit two key weaknesses:
(1) they focus on the limited setting of image captioning, and none of them explored evaluating hallucination of the complex and free-form answers to the open-ended questions~\cite{OpenAI2023GPT4V} (e.g. multimodal conversations~\cite{llava,DBLP:conf/acl-convai/SundarH22}, world knowledge-based VQA~\cite{AOKVQA} and visual storytelling~\cite{Huang2016VisualStory});
(2) they ignore fine-grained hallucinations of visual attributes in the generated answer \cite{liu2024mitigating,DBLP:conf/emnlp/RohrbachHBDS18}. 

Evaluating hallucinations present in free-form answer is especially challenging for two primary reasons:
(1) \textbf{Free-form answers contain a hybrid of descriptive and analytical contents.} Unlike close-domain tasks such as image captioning, answering open-domain questions in a free-form manner does not only require generating the question-relevant descriptive content of the given image. It may also involve analytical content such as rationales that include external commonsense knowledge. As depicted in Figure~\ref{fig:hallucination}, certain sub-sentences (\ie those without the underline) do not require verification with the image input due to their analytical nature. 
Because they encompass subjective analytical content that extends beyond a direct description of the visual inputs.
Neglecting to distinguish between analytical and descriptive content inevitably distracts the factual measurement.
Thus, pinpointing the descriptive content within the answers generated by LVLMs is significant. 
(2) \textbf{Model outputs are prone to the multiplicity of hallucinations.} Current methodologies offer a constricted view on evaluating hallucinations, primarily concentrating on coarse-grained object existences~\cite{DBLP:conf/emnlp/RohrbachHBDS18,lovenia2023negative}, while neglecting other fine-grained elements, such as counts, colors, and the interrelations between objects (\eg the spatial relation between the person and the car in Figure~\ref{fig:hallucination}), which also form a significant portion of visual hallucinations~\cite{gunjal2023detecting}.
Consequently, devising a method to holistically evaluate fine-grained hallucinations of visual attributes is also important.





To address the aforementioned challenges, we propose the \ours\ metric, which can evaluate \textit{fine-grained hallucinations in various multimodal tasks}, such as image captioning and open-ended questions. This metric comprises three primary components: Descriptive Sub-sentence Identification, Atomic Fact Generation, and Fact Verification, as illustrated in Figure~\ref{fig:ours}. The first component is tasked with discerning descriptive sub-sentences within the composite content of the generated free-form answer. Thereafter, the second component deconstructs this descriptive content into fine-grained elements (\ie atomic facts)~\cite{factscore}. These atomic facts cover comprehensive types, such as objects attributes and interrelationships. The last component emphasizes verifying the consistency between the visual information and the derived atomic facts via a Visual Entailment Model (VEM)~\cite{DBLP:journals/corr/abs-1901-06706}. Based on the proposed metric, we evaluated several advanced LVLMs, such as LLaVA~\cite{llava} and MiniGPT-4~\cite{minigpt4}. From the results, we conclude that current LVLMs still face challenges of answers that are not faithful to the input image, which leaves a large room for improvement.

In summary, our contributions are as follows: 
(1) We introduce \ours, a metric tailored to assess hallucinations in LVLMs free-form answers to open-ended questions, which is not yet addressed by current studies; 
(2) To the best of our knowledge, we are the first to systematically study the LVLMs free-form answers and evaluate the fine-grained hallucinations of various types in LVLMs; 
(3) In our quest to understand the hallucinations manifested by LVLMs, we embark on comprehensive experiments with six open source models across diverse tasks and datasets. Our findings underscore that the hallucination phenomenon remains a pressing challenge for current LVLMs. As a byproduct, we released our code\footnote{\url{https://github.com/bcdnlp/FAITHSCORE}.}.

%% file: sec/2_related_work.tex
\section{Related Work}
\paragraph{Large Vision-Language Model}
Motivated by the success of the pretraining technique in LLMs and VFMs, the multimodal learning research community has recently shifted the research attention to LVLMs~\cite{DBLP:journals/corr/abs-2308-01390,DBLP:journals/corr/abs-2305-03726}. 
Contemporary advanced LVLMs predominantly feature three core components: a text encoder, an image encoder, and a cross-modal alignment module~\cite{DBLP:conf/emnlp/RohrbachHBDS18}. Specifically, the text encoder often takes the form of a language model, as seen in examples like LLaMA~\cite{llama} and Vicuna~\cite{vicuna}. Conversely, the image encoder is typically derived from VFMs, such as ViT~\cite{vit}. The function of the cross-modal alignment module is to bridge visual content with textual representation, enhancing the text encoder's capacity to interpret visual semantics. 
To accomplish visual understanding, LVLMs typically undergo multiple training phases~\cite{MultiModal-GPT,minigpt4,llava15,llava,mPLUG-Owl,InstructBLIP}. 
For instance, \citet{llava} first aligns the image features with the word embeddings of a pre-trained LLM during an initial pre-training stage, and subsequently fine-tunes the LVLM using specialized language-image instruction-following datasets. 
For efficiency enhancement, LVLMs often freeze parameters of the LLM or VFM and are trained with efficient fine-tuned techniques~\cite{mPLUG-Owl,InstructBLIP}, such as LoRA~\cite{lora}.

However, in spite of the considerable advancements made by LVLMs, they consistently grapple with hallucination issues. These issues markedly impact their efficacy across a range of vision-language tasks~\cite{DBLP:conf/emnlp/RohrbachHBDS18}. 



\paragraph{Vision-language Model Hallucinations and Evaluations}
Though hallucination phenomenons and mitigation methods have been extensively studied in the text generation literature~\cite{ji2023survey, factscore}, it is much less investigated in vision-language models~\cite{InstructBLIP, llava,fgaif,mmss}.
Although there are a few existing works tackling this issue, they mainly focus on the constraint problem setting such as image captioning~\cite{densecap}.
For example, 
\citet{DBLP:conf/emnlp/RohrbachHBDS18} propose caption hallucination assessment with image relevance (CHAIR), which is a popular metric for evaluating object hallucination in sentence-level captions. They also show that popular metrics like METEOR~\cite{meteor} and CIDEr~\cite{vedantam2015cider} do not capture this. 
\citet{Li2023EvaluatingOH} extends CHAIR and proposes ``POPE'', a polling-based query technique for probing objects. Besides, \citet{lovenia2023negative} devised Negative Object Presence Evaluation (NOPE) to quantitatively assess object hallucination through VQA, based on ``POPE''. 
\citet{gunjal2023detecting} further proposed to detect hallucinations in more detailed image captions and investigated utilizing a reward model for mitigating them. \citet{DBLP:journals/corr/abs-2309-04041} introduced an evaluation benchmark that contains more diverse types of questions, such as Yes-or-No and Fill-in-the-Blank. 

Different from all the above, we are the first to propose a \textcolor[rgb]{0,0,0}{reference-free} metric for \textcolor[rgb]{0,0,0}{fine-grained} evaluating the answers in the \textcolor[rgb]{0,0,0}{open-ended visual question-answering setting}, where answers are of free form and can be lengthy.

%% file: sec/3_method.tex
\section{Estimating \ours}
\label{sec:method}
In this section, we begin by clearly defining the research problem in Section \ref{31}, followed by a detailed framework of estimating \ours\ in Section \ref{32}.

 \begin{figure}[t]
\centering \footnotesize
\includegraphics[width=\linewidth]{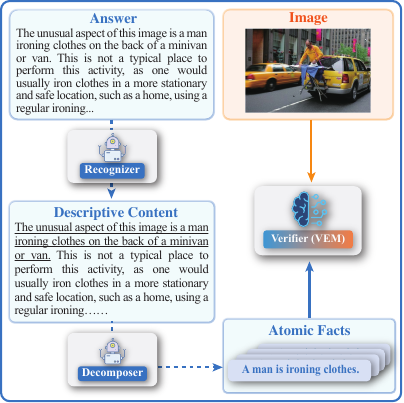}
\caption{
    An overview of estimating \ourstitle, which mainly consists of three steps: Descriptive Sub-sentence Identification, Atomic Fact Generation, and Fact Verification. These steps are implemented by three modules: Recognizer, Decomposer, and Verifier. The \underline{underlined part} denotes recognized descriptive content.
}\label{fig:ours}
\vspace{-2mm}
\end{figure}

\subsection{Task and Settings}
\label{31}
Suppose we have an image $I$ and a corresponding prompt $Q$. We then feed them into the LVLM denoted as $\mathcal{M}$, to obtain the generated answer $A$. 
Our objective is to design a scoring framework to estimate \ours\ $f$ based on the input prompt $Q$, the input image $I$, and the generated answer $A$. 
It is defined as:
$s = \mathcal{F} (A, Q, I)$.  
$s$ is a scalar value ranging between 0.0 and 1.0. Notably, the devised evaluation method is reference-free and doesn't require a ground truth answer. 

\subsection{The Evaluation Framework}
\label{32}
In order to estimate \ours of the generated answers, we introduce a novel framework to implement the scoring function $\mathcal{F}$. The framework comprises three key steps: descriptive sub-sentence identification, atomic fact generation, and fact verification, as depicted in Figure~\ref{fig:ours}. 

\paragraph{Descriptive Sub-sentence Identification.} 
Faithfulness in the context of LVLMs refers to the consistency between the input visual content and the generated answer. Notably, we focus on the details in the answer that describe the input image objectively, to obtain a more precise and fine-grained understanding of the hallucination. 
As shown in Figure~\ref{fig:hallucination}, only some sub-sentences (\ie those with the underline) from the answer require verification with the image input.
 Hence, we need to identify the descriptive sub-sentences from the answer using a recognizer. The sub-sentences denote the short sentences that are split by punctuation in the answer. 

Humans are capable of distinguishing descriptive sub-sentences from other contents (referred to as analytical sub-sentences) by analyzing the semantics of the answers generated by LVLMs. However, manually identifying descriptive sub-sentences is a resource-intensive process, requiring plenty of human labor. To address this problem, we turn to ChatGPT to implement the recognizer as a practical solution, since it has demonstrated 
remarkable semantics understanding capabilities across a wide range of natural language processing tasks~\cite{chatgpt}. Section~\ref{sec:42} shows that ChatGPT can achieve promising performance on this task.

To be more specific, our approach first crafts a prompt $P$ that encompasses task instructions and $K_1$ in-context learning examples. We feed this designed prompt along with the to-be-processed answer $A$ into the ChatGPT, generating the recognized results, defined by the equation 
$\hat{A} = ChatGPT(A, P)$, 
where $\hat{A}=\{\{a_1,l_1\},\cdots,\{a_k,l_k\}\}$ signifies the generated result, in which the answer is split into sun-sentences $a$, and every sub-sentence is assigned a label $l$ (\ie descriptive or analytical). Then we extract all descriptive sub-sentences denoted as $A'=\{a'_1,\cdots,a'_t\}$. For a more comprehensive understanding of the specific prompt $P$ utilized in this process, please refer to Section~\ref{prompts} of the Appendix.

         

\paragraph{Atomic Fact Generation.} 
Despite we have identified descriptive sub-sentences from the answer, there are still multiple facts hybrid in each sub-sentence.
Each descriptive sub-sentence consists of multiple pieces of information (\ie atomic facts), each of which may contain hallucination. Therefore, to access a fine-grained evaluation, we design a decomposer to further break the sub-sentences into atomic facts. 
In particular, 
we define atomic facts as an element belonging to an entity, relation, or attribute, inspired by the existing works~\cite{factscore,TIFA}. 
Importantly, the atomic fact is a minimal unit of information. This handling can ensure the verification of \emph{each} element in the answer without being disturbed by other information. 
Atomic facts include three types: entity existence, attributes, and relations. An entity fact indicates an object’s existence. Attribute facts relate to characteristics like color and shape. Relation facts describe inter-entity relationships, e.g., the spatial relation. 
In Figure~\ref{fig:hallucination}, we show some examples of atomic facts. 


To achieve this, similar to the process of identifying descriptive sub-sentences, we also utilize the ChatGPT for the generation of atomic facts. This is because ChatGPT has shown a strong ability in information extraction~\cite{DBLP:journals/corr/abs-2302-10205}. More precisely, we annotate a set of $K_2$ examples for demonstrations and prompt the ChatGPT for atomic fact generation with $P'$ as follows:
${E_i}= ChatGPT (A', P'), i \in [1,C]$, 
where $A'$ are all descriptive sentences identified in step 1, 
${E_i}=\{e_i^1,\cdots,e_i^{n_i}\}$ represents all $n_i$ atomic facts belonging to the $i$-th category, and $C$ stands for the total number of categories (\ie $C=5$ in our case) \textcolor[rgb]{0,0,0}{and the category include ``entity'', ``relation'', ``color'', ``count'', and ``others''}. 
Further details regarding the specific prompt $P'$ utilized in this process can be found in Section~\ref{prompts} of the Appendix.

\paragraph{Fact Verification.}
In this stage, we compare the atomic facts derived above with the image to determine if the facts are faithful to the input visual information. Specifically, to calculate the \ours\ for the derived atomic facts, we first compute the score for each fact and then aggregate them to derive the overall score using the following formula:
\begin{equation}
\hat{s} = \frac{\sum_{i=1}^{C} \sum_{j=1}^{n_i} {w_i^j \cdot s (e_i^j, I )} }{ \sum_{i=1}^{C} {n_i} },
\end{equation}
where $\hat{s}$ represents the overall \ours\ of the answer $A$. The function $s(e_i^j, I)$ refers to the verification function (\ie Verifier), which measures whether $e_i^j$ can be supported by the input image $I$. The parameter $w_i^j$ is a weighted factor that can be used to assign different weights to different atomic facts for various tasks. 
To implement function $s(e_i^j, I)$, we resort to the Visual Entailment Model (VEM) (\eg OFA~\cite{ofa}), which is able to predict whether the image semantically entails the text. We elaborate on selections of the verifier models in Section \ref{43verifer}.
In particular, when the output of the VEM is positive, indicating that the image $I$ semantically entails the text $e_i^j$ resulting in $s(e,I)=1$, and negative otherwise. 
In this work, we set all the weights $w_i^j$ to 1, following the setting of the existing work~\cite{factscore,DBLP:conf/eacl/KrishnaBKIDCL23}. 
In addition, we further introduce a sentence-level \ours\ metric as follows, $\hat{s}_s = 1 - S_h/{S}$, 
where $S$ is the total number of descriptive sub-sentences in the answer and $S_h$ is the total number of descriptive sub-sentences with hallucinations.

%% file: sec/4_meta.tex
\section{Meta-evaluate \ours\ }
\label{sec:meta}


To verify that our automatic evaluation correlates with human judgment, 
we conduct human evaluations in terms of hallucination. We select the test dataset from the LLaVA paper~\cite{llava}  (LLAVA-Bench) for human evaluation, which is constructed based on the MSCOCO dataset.  
This test set is a visual instruction following dataset comprising three distinct question types: detailed description, conversation, and complex question. For each type, this dataset includes 90 questions. We select answers from  LLaVA~\cite{llava} and InstructBLIP~\cite{InstructBLIP} models for evaluation. 


\subsection{Human Evaluations of Hallucinations}

For each test example, we craft an annotation process to assign the faithfulness score to models' generated answers via the subsequent steps.

\vspace{.5em}
\noindent \textbf{Step 1: Sub-sentence Identification.}
Annotators first review the given question, the corresponding answer, and the associated image. Subsequently, they evaluate each sub-sentence extracted from the answer. If a sub-sentence is an objective description of visual information, they mark it as the ``descriptive'' category; otherwise, it's categorized as ``analytical''. For the  ``analytical'' sub-sentence, annotators should skip the following steps. Otherwise, they should follow the next steps.

\begin{table}[t]
    \centering
    \resizebox{\columnwidth}{!}{\begin{tabular}{lccc} 
    \toprule 
           \multirow{2}*{\bf Recognizer} & \multicolumn{2}{c}{\bf LVLM} & \multirow{2}*{\bf Overall}\\

     ~&  \bf LLaVA &\bf InstructBLIP  & \\ \midrule

        ChatGPT &  89.84&  91.58 &90.74 \\
        LLaMA-7B &  68.01& 71.39& 69.75 \\
        LLaMA-7B (w/ context) & 72.80 & 66.76& 69.68 \\
        \bottomrule
        
    \end{tabular}}
    \caption{Comparison of recognizer LLMs' accuracy (\%) on identifying descriptive sub-sentences. 
    For LLaMA, we used two different prompt settings, either to input only the sub-sentence or both the sub-sentence and its context into the model (LLaMA-7B w/ context).
    }
    \label{tab:identifier}
    
\end{table}

\vspace{.5em}
\noindent \textbf{Step 2: Atomic Fact Generation and Revision.}
In this step, human annotators are asked to decompose the descriptive sub-sentences into a sequence of atomic facts. To optimize the annotation process and reduce the time required, we pre-supply atomic facts derived from ChatGPT. Annotators then have the flexibility to use or modify these facts as needed. In particular, annotators examine each atomic fact to ensure its fidelity to the given sub-sentence. The facts that are either redundant or non-atomic are asked to be removed. Subsequently, the focus shifts to the linguistic aspect, ensuring that each atomic fact is articulated in a coherent manner and that it accurately represents the original entity or concept of the answer by revising facts manually. Additionally, any missing atomic facts from the descriptive sub-sentence are added. 
For the process of removing and revising atomic facts, please refer to the Interface functionalities in the Appendix. \textcolor[rgb]{0,0,0}{ Errors introduced by the ChatGPT in this stage are shown in Appendix \ref{app:decompose}.}

\vspace{.5em}
\noindent \textbf{Step 3: Fact Verification.}
In this step, for every individual atomic fact derived from the descriptive sub-sentence, annotators assess its consistency with the given image. If the content of atomic facts is not present or contradicts the image, it's identified as a hallucination, and accordingly marked as ``yes''. Conversely, if the element is in alignment with the image, it's validated and marked as ``no''. 
To quantify the human evaluation of faithfulness, we employ the Likert Scale~\cite{likert1932technique}. This approach transforms human evaluations into a tangible scale, ranging from 1 (being the poorest) to 5 (being the best). The details about the annotation process are given in Section~\ref{appenlikert} of the Appendix.





\begin{table}[t]
    \centering
    \resizebox{\columnwidth}{!}{
    \begin{tabular}{lccc} 
    \toprule 
           \multirow{2}*{\bf Verifier} & \multicolumn{2}{c}{\bf LVLM} & \multirow{2}*{\bf Overall}\\

     ~&  \bf LLaVA &\bf InstructBLIP  & \\ \midrule
        OFA-EM &  81.07&  78.08& 79.42\\
        OFA &  84.47& 80.71& 82.39\\
        mPLUG&   84.95& 83.86& 84.35\\
        BLIP-2-flant5xl&  78.64& 77.42&77.97\\ 
        BLIP-2-flant5xxl&  82.36& 83.20& 82.83\\
        LLaVA&  67.25& 67.10& 67.17\\ 
        LLaVA-1.5& 85.65& 84.49& 85.07 \\ 
        \bottomrule
    \end{tabular}
    }
    \caption{Comparison of the Verifier LLMs accuracy on verifying the atomic facts (the third step).}
    \label{tab:vems}
    \vspace{-4mm}
\end{table}

\subsection{Recognizer Accuracy on Descriptive Sub-sentence Identification}
\label{sec:42}
To obtain the performance of recognizers (e.g. LLMs) on the sub-sentence identification task, we construct a sub-sentence identification dataset based on our annotated samples. The final label for each sub-sentence is determined by the majority voting scheme. The total number of sub-sentences is $1,382$ and the average number of sub-sentences in the answer is $7.68$. 
We select the superior ChatGPT (Proprietary) and LLaMA-7B (Public) models for this task and report their accuracy on identifying descriptive sub-sentences. The results are shown in Table~\ref{tab:identifier}. ChatGPT outperforms LLaMA-7B on sub-sentence identification. 
For LLaMA-7B based method, when additional context beyond the sub-sentence itself is included, there is an improvement on LLaVA answers test set, but overall there is no significant improvement.


\subsection{Verifier Accuracy on Fact Verification}
\label{43verifer}
Another key factor of our automatic method is the reliability of the verifier visual entailment model (VEM). Hence, we also evaluate the accuracy of different VEMs on the annotated samples. Because of the atomic fact revision operation during the annotation process, there may be some differences in atomic facts labeled by different annotators. To improve reliability, we only keep these atomic facts annotated by all three annotators for VEM evaluation. The final label for each atomic fact is determined by the majority voting scheme. The total number of atomic facts derived from descriptive sub-sentences is $1,380$ and the average number of atomic facts in each descriptive sub-sentence is $2.04$. For verifier VEMs, we evaluate OFA-EM, OFA~\cite{ofa}, mPLUG~\cite{mplug}, BLIP-2-flant5xl, BLIP-2-flant5xxl~\cite{blip2}, LLaVA, LLaVA-1.5 
(Table~\ref{tab:vems}). More details about these models are shown in Section~\ref{devem} of the Appendix. 
Among all models, LLaVA-1.5 performs best on fact verification, so we use it for estimating \ours\ in Section 5.  
\textcolor[rgb]{0,0,0}{
Another potential reason for employing this LVLM as a verifier is that our verification task is a discriminative task that usually generates a shorter response, and tends not to generate hallucination, which has been demonstrated in our Section 5.5 (The Influence of Answer Length on Hallucinations) and the existing work \cite{factscore,TIFA}.
}


\begin{table}[]
    \centering \small
    \begin{tabular}{llll}
    \toprule
    \bf Metric & $r$ (\%) & $\rho$ (\%) & $\tau$ (\%) \\ \midrule
        BLEU-4 & -1.9& -8.2 & -5.8 \\ 
        ROUGE-L & -8.7&-6.2 &  -4.7\\ 
        METEOR & -12.2&-8.5 & -6.3\\ 
        CHAIR & 16.8&19.2 & 14.8\\ 
        CLIP-Score &19.8 & 16.6  & 11.7 \\  
SPICE& 20.2 &21.3 &25.4\\ 
\midrule

        Ours & 48.2&  38.4& 47.6\\ 

    \bottomrule
    \end{tabular}
    \caption{Correlation between each evaluation metric and human judgment on LVLMs \textcolor[rgb]{0,0,0}{(i.e., LLaVA and InstructBLIP)}
    hallucinations, measured by Pearson's $r$, Spearman’s $\rho$, and Kendall’s $\tau$. 
    \textcolor[rgb]{0,0,0}{The p-value of the significant test between our result and the baseline result is less than 0.01.}
    } 
    \label{tab:correlation}
\end{table}

\subsection{Correlations with Human Evaluations}



To prove the superiority of our proposed metric \ours, we compare it with several multimodal generation evaluation metrics: 1) reference-based: BLEU-\{4\}~\cite{bleu}, Rouge-\{L\}~\cite{ROUGE}, METEOR~\cite{meteor}, CHAIR~\cite{DBLP:conf/emnlp/RohrbachHBDS18}, SPICE~\cite{DBLP:conf/eccv/AndersonFJG16} and 2) reference-free: CLIP-Score~\cite{clipscore}. 
Table~\ref{tab:correlation} delineates the correlation between various evaluation metrics and human judgment regarding LVLM faithfulness.
\textcolor[rgb]{0,0,0}
 {Among all metrics, our metric \ours\ achieved the best correlation with human correlation. More details and analysis about human correlation can be found in Appendix \ref{sec:human_eval} and \ref{app:cost}.}





%% file: sec/5_atomic_eval.tex
\section{Evaluating Vision-Language Model Hallucinations with  \ours}

\subsection{Models and Datasets}
We selected six open-source LVLMs for evaluation.
1) MiniGPT-4~\cite{minigpt4};
2) LLaVA~\cite{llava}; 
3) InstrucBLIP~\cite{InstructBLIP}; 
4) Multimodal-GPT~\cite{MultiModal-GPT}; 
5) mPLUG-Owl~\cite{mPLUG-Owl}; 
6) LLaVA-1.5.


To assess the performance of existing LVLMs, we conducted experiments using two datasets. Here is a description of each dataset:
(1) MSCOCO-Cap: This dataset is designed for the image captioning task. We randomly select 1,000 images from the MSCOCO~\cite{mscoco} validation set and devised the prompt as ``Generate a concise caption for the given image''; 
(2) LLaVA-1k: We extract 1,000 images from the MSCOCO validation set and generated three types of prompt-answer pairs (\ie detailed description, conversation, and complex question) for each image by ChatGPT, following the data generation method in~\cite{llava}.

\begin{table*}[t]
    \centering \small
    \begin{tabular}{lccccc}
        \toprule
            \bf  &  \multicolumn{4}{c}{\bf LLaVA-1k} & \bf MSCOCO-Cap \\
         \bf  & \bf Conversation & \bf Detailed Description & \bf Complex Question & \bf Overall & -\\
        \midrule
            Multimodal-GPT& 0.5321&0.5299&0.5385&0.5335 & 0.5440\\

         MiniGPT-4 & 0.5679& 0.5768 & 0.5691 & 0.5713 & 0.6359\\
        mPLUG-Owl  & 0.7246 & 0.7240 & 0.7015& 0.7167 & 0.8546\\
         InstructBLIP & 0.8061& 0.8161& 0.8049& 0.8091 & {0.9392}\\

                  LLaVA & 0.8302& 0.8386&0.8392 & 0.8360& 0.8729\\
        LLaVA-1.5& \bf{ 0.8569}&\bf{ 0.8611}&\bf{ 0.8516} &  \bf{0.8566} & \bf 0.9425\\
        

        \bottomrule
    \end{tabular}
    \caption{\ours\ evaluation results ($\uparrow$) of different LVLMs on the LLaVA-1k and MSCOCO-Cap datasets.}
    \label{tab:llavadataset1k}
\end{table*}

\begin{table*}[t]
    \centering \small
    \begin{tabular}{lccccc} \toprule
    \bf  &  \multicolumn{4}{c}{\bf LLaVA-1k}  & \bf MSCOCO-Cap\\
\bf         & \bf Conversation&\bf Detailed Description &\bf Complex Question &\bf Overall & -\\ \midrule
         Multimodal-GPT& 0.4615&0.4827&0.5131&0.4858 & 0.6277\\
         MiniGPT-4 & 0.6441& 0.6489 & 0.6499 & 0.6476 &0.6017\\
                           LLaVA & 0.7106 & 0.6979&0.7038& 0.7041 & 0.6681\\
                           InstructBLIP & 0.7231 & 0.7327 & 0.7149& 0.7236 & {0.7970}\\
                  mPLUG-Owl & 0.7369& 0.7163& 0.7344& 0.7292 & 0.6447\\ 

        LLaVA-1.5 &\bf 0.7722 &\bf 0.7717&\bf 0.7699&\bf 0.7713 &\bf 0.8258\\

        \bottomrule
    \end{tabular}
    \caption{\ours\ (sentence-level) evaluation results ($\uparrow$) of different LVLMs.
    }
    \label{tab:sentence1}
\end{table*}



\subsection{Hallucination Evaluation}

\begin{figure}
    \centering
    \includegraphics[width=\linewidth]{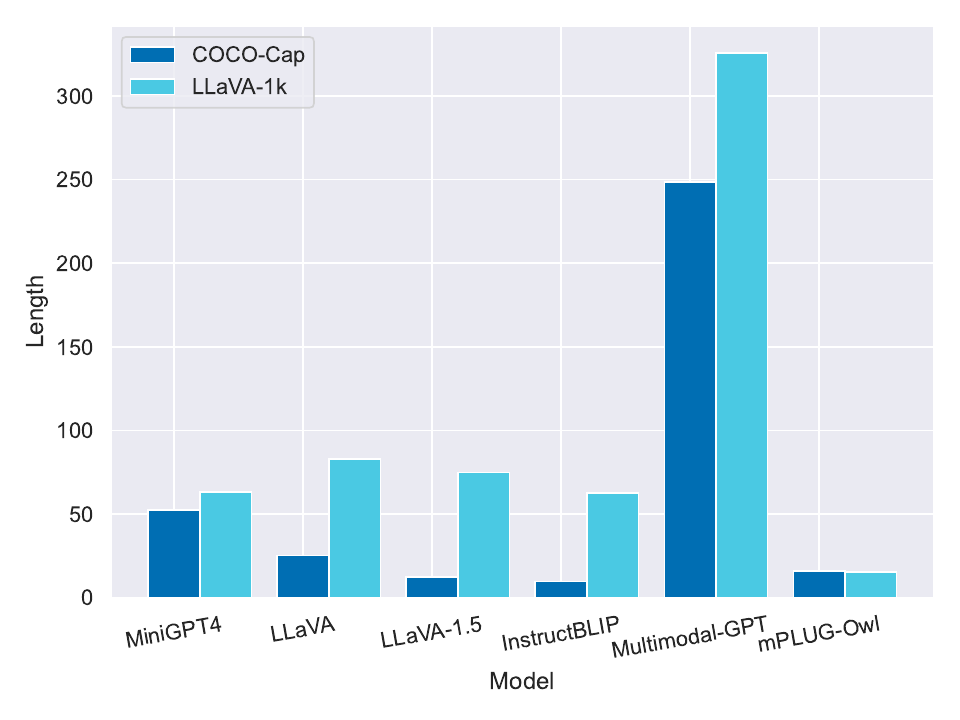}
    \caption{Answer lengths distributions of different models on two benchmark datasets.}
    \label{fig:length}
    \vspace{-4mm}
\end{figure}

Table~\ref{tab:llavadataset1k} 
presents a comprehensive performance comparison of various models in terms of \ours\ when benchmarked on the LLaVA-1k and MSCOCO-Cap datasets. We observe that: 
(1) LLaVA-1.5 outperforms their counterparts in most situations. This demonstrates their preeminent capability in achieving and maintaining faithfulness during generation processes. 
(2) It's worth noting that different models have similar performance across tasks. For instance, MiniGPT achieved 0.5679,  0.5768, and 0.5691 \ours\ on the ``Conversation'', ``Detailed Description'', and ``Complex Question'' tasks, respectively. 
(3) For most models, the performance on the MSCOCO-Cap dataset is better than their performance on the LLaVA-1K dataset. The potential reason may be that model answers to the MSCOCO-Cap questions are usually shorter than their answers to the LLaVA-1K questions (see Figure~\ref{fig:length}).

\subsection{Sentence-level Hallucination Evaluation}
To further understand the faithfulness of LVLMs, we evaluate them with the \ours\ (sentence-level). Table~\ref{tab:sentence1} shows the sentence-level \ours\ evaluation across different LVLMs. 
Multimodal-GPT achieves poor performance in \ours\, it also performs less favorably in terms of sentence-level hallucination evaluation. In addition, LLaVA-1.5 performs well in terms of \ours\ and \ours\ (sentence-level). This indicates the consistency between \ours\ and sentence-level \ours.

\begin{figure}[t]
    \centering
    \includegraphics[width=\linewidth]{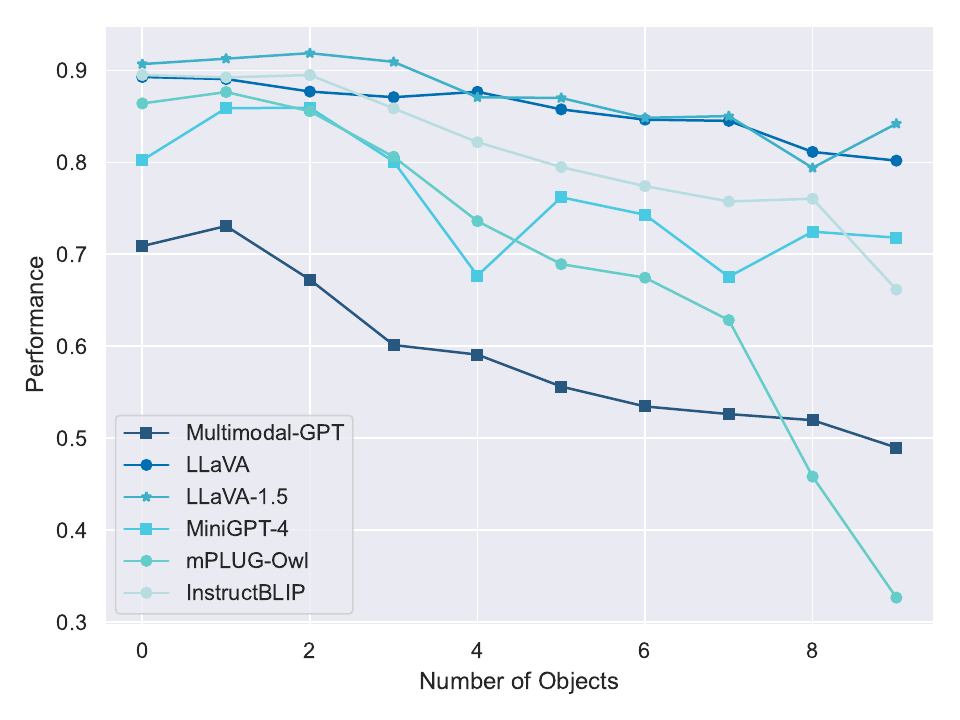}
    \caption{The relation between \ours\ and numbers of objects (\ie entities) in the answers (LLaVA-1k dataset). As the number of entities increases, model performance (\ie \ours) drops significantly. 
}
\vspace{-4mm}
    \label{fig:objects}
\end{figure}

\subsection{Other Analysis}
\paragraph{The Influence of Answer Length on Hallucinations.} 
To further elucidate the impact of answer length on hallucinations, we analyze answer lengths across various LVLMs on different datasets. As illustrated in Figure~\ref{fig:length}, there's a significant variation in the distribution of answer lengths produced by different models. Multimodal GPT consistently generates the lengthiest responses, potentially compromising its performance across tasks. In contrast, mPLUG-Owl tends to produce shorter answers than its counterparts, hence it may generate fewer hallucinations. 
Meanwhile, the image captioning task showed better faithfulness in generated content than the other task for most LVLMs. This may be attributed to the fact that captioning sentences mainly are brief descriptions and shorter. 

\begin{figure}[t]
    \centering
    \vspace{-4mm}
\includegraphics[width=\linewidth]{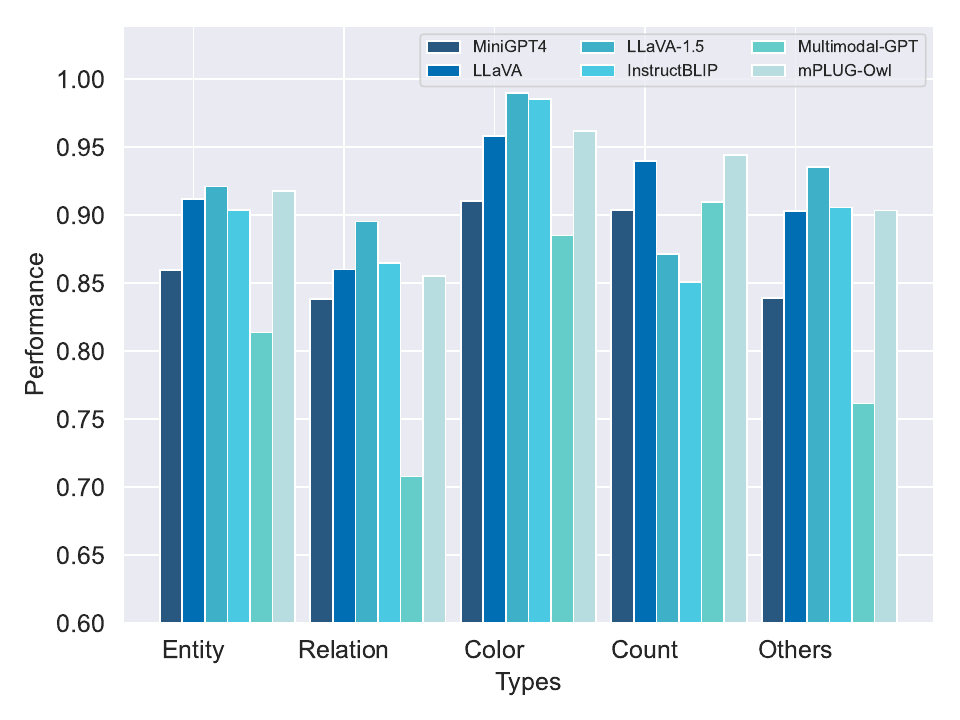}
    \caption{\ours\ on each type of atomic facts on the LLaVA-1k benchmark. The types are \textsc{Entity}, \textsc{Relation}, \textsc{Color}, \textsc{Count}, and \textsc{Others}.
    }
    \label{fig:types}
    \vspace{-4mm}

\end{figure}

\paragraph{The Influence of Multiple Objects.}
Figure~\ref{fig:objects} shows how the number of objects in the answer generated by different models affects the  \ours\. 
The model's faithfulness varies with the number of objects. While all models start with relatively high scores when there are few objects in the answer, their performance generally drop as the number of objects increases. For example, Instruct-BLIP starts with a high \ours\ of 0.895 for 1 object and sustains a relatively low score of 0.662 for 10 objects.


\paragraph{Analysis on Types of Hallucination}
To deduce the model strengths and vulnerabilities of each in maintaining faithfulness, we compared the faithfulness performance of various models across different categories of hallucination.
We mainly investigated the five distinct categories: \textsc{entity}, \textsc{count}, \textsc{color}, \textsc{relation}, and \textsc{other} attributes, motivated by the existing works. 
From Figure~\ref{fig:types}, we can observe that while LLaVA-1.5 consistently excels across most categories, other models also showcase strengths in specific domains. 
The bad performance of some types may provide insightful information for model improvement. 
Importantly, achieving consistently high faithfulness across a diverse range of categories remains a formidable challenge for LVLMs. 

%% file: sec/6_conclusion.tex
\section{Conclusion}
We introduce a novel metric called \ours\ for evaluating free-form and open-domain answers generated by large vision-language models. Compared to previous metrics, \ours\ offers a finer level of granularity, interpretability, and closer alignment with human judgments. Our quantitative analysis demonstrates that current LVLMs are prone to visual hallucination problems. We also find that the answer length and number of objects could affect the faithfulness of LVLMs. In addition, the faithfulness performance of LVLMs on different types of atomic facts varies. 
We expect that \ours\ will be of great value for evaluating forthcoming advanced LVLMs. 


%% file: sec/X_suppl.tex


\clearpage
\appendix


\section{Likert Scale Guideline} \label{appenlikert}
For human evaluation, we utilize the Likert Scale to get the final faithfulness score for each testing sample.  
Specifically, suppose the generated answer consists of $n$ atomic facts, out of which $x$ atomic facts are determined as hallucinations. Both $n$ and $x$ are labeled by the annotators. The benchmark scoring guideline is outlined as follows:
\begin{itemize}
    \item Score 1: All atomic facts are hallucinations, symbolized as $x == n$;
    \item Score 2:  More than half of the atomic facts are hallucinations, represented as $x>n/2$;
    \item Score 3: Half or fewer atomic facts are hallucinations, represented as $n/3<=x<n/2$;
    \item Score 4: Less than one-third of the atomic facts are hallucinations, which translates to $x<n/3$;
    \item Score 5: All atomic facts accurately represent the visual content, meaning $x=0$.
\end{itemize}

         


\section{Details about VEMs}
\label{devem}
We select OFA-EM, OFA\footnote{https://github.com/OFA-Sys/OFA.}~\cite{ofa}, mPLUG\footnote{https://github.com/X-PLUG/mPLUG.}~\cite{mplug}, BLIP-2-flant5xl, BLIP-2-flant5xxl\footnote{https://github.com/salesforce/LAVIS/tree/main/projects/blip2.}~\cite{blip2}, LLaVA~\cite{llava}, and LLaVA-1.5\footnote{https://github.com/haotian-liu/LLaVA.}~\cite{llava15} as VEM and evaluate them based on our annotated dataset.. OFA-EM is an open-source model which was finetuned on the SNLE-VE dataset~\cite{DBLP:journals/corr/abs-1901-06706}. Hence, this model can tackle visual entailment tasks directly. For the OFA-EM model, the ``neutral’’ is categorized as hallucination because the OFA can't decide whether the verified content appears in the input image. 
For the other models, they are also open-source and finetuned on the visual question answering dataset. To enable them to tackle the visual entailment task, we get an input a prompt ``Statement: \{\textit{atomic facts}\} Is this statement right according to the image? Please output yes or no.'', into models.

\section{Testing Examples of GPT-4Vision} 
\label{exgpt4v}

\paragraph{Hallucination in Advanced GPT-4Vision}

\begin{figure*}[t]
    \centering
    \includegraphics[width=0.95\linewidth]{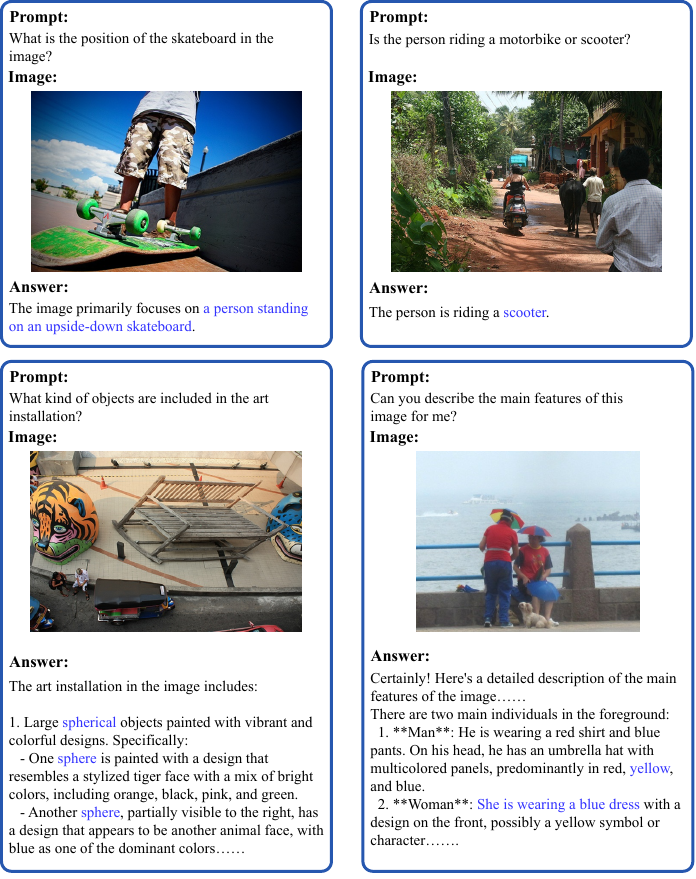}
    \caption{Illustration of answers generated from GPT-4Vision. Similar to existing open-source VLMs, GPT-4Vision is still prone to the hallucination problem in the generated answer. The \textcolor{blue}{blue contents} denote hallucinations.}
    \label{fig:samples-gpt-4}
\end{figure*}

Here we test the GPT-4Vision model on four examples. Based on the results, we can come to the conclusion that the GPT-4Vision answers still contain various hallucinations despite it may have very large parameters and have been trained with a large corpus, as shown in Figure~\ref{fig:samples-gpt-4}.

\section{More benchmarks}
We further compute our metric on one dataset: LRV-Instruction~\cite{DBLP:journals/corr/abs-2306-14565}. The results are shown as follows, which are consistent with the experimental results on datasets LLaVA-1k and MSCOCO-Cap:  
InstructBLIP 0.6626, 
MultimodalGPT 0.4903, 
mPLUG-Owl 0.6433, 
MiniGPT-4 0.4638, 
LLaVA 0.7017, 
LLaVA-1.5 0.7855.

\section{Examples of Evaluation}
\label{exanno}
\begin{figure}[t]
    \centering
    \includegraphics[width=\linewidth]{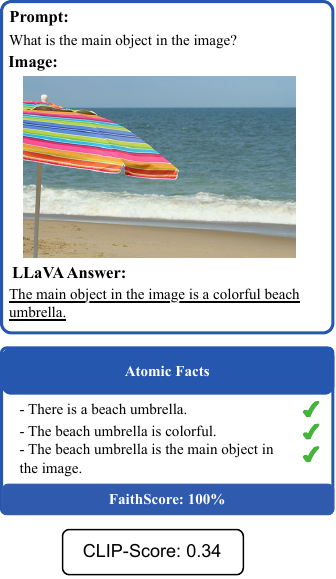}
    \caption{Illustration of how \ours\ is computed for a testing sample.  The underlined contents denote recognized descriptive content.}
    \label{fig:samples1}
\end{figure}

\begin{figure}[t]
    \centering
    \includegraphics[width=0.85\linewidth]{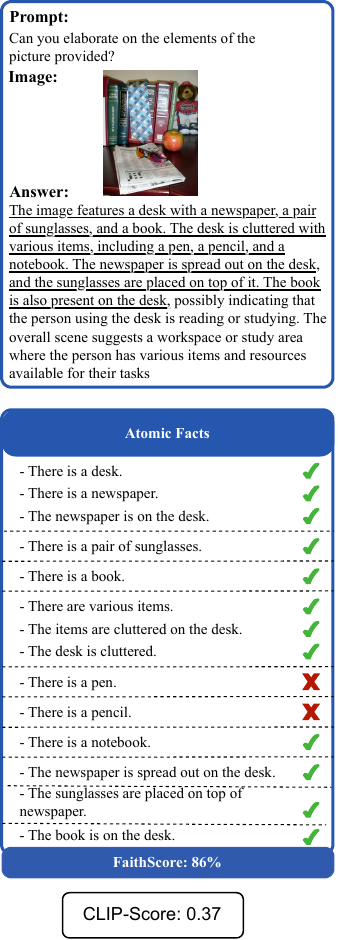}
    \caption{Illustration of how \ours\ is computed for a testing sample. The underlined contents denote recognized descriptive content.}
    \label{fig:samples2}
\end{figure}

\begin{figure}[t]
    \centering
    \includegraphics[width=0.85\linewidth]{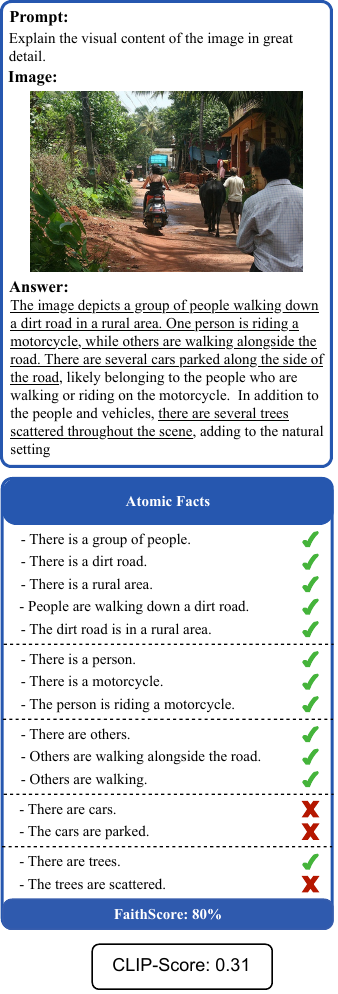}
    \caption{Illustration of how \ours\ is computed for a testing sample. The contents highlighted by the underline denote recognized descriptive content.}
    \label{fig:samples3}
\end{figure}

\begin{figure}[t]
    \centering
    \includegraphics[width=\linewidth]{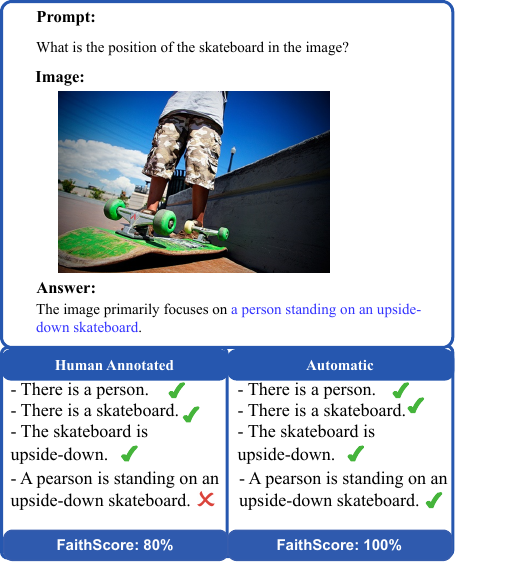}
    \caption{Illustration of the case that the proposed metric score disagrees with human judgement}
    \label{fig:sampleserror}
\end{figure}

Here we show three examples of how \ours\ is computed and the existing best reference-free CLIP-Score value in Figure~\ref{fig:samples1}, Figure~\ref{fig:samples2}, and Figure~\ref{fig:samples3}. 
Additionally, we present an example (see Figure~\ref{fig:sampleserror}) where the proposed metric score diverges from human judgment, illustrating a discrepancy attributed to an error generated by the recognizer system.

\section{More Details about Human Evaluation}
\label{sec:human_eval}

We employ 3 workers for annotation and each person annotated 180 testing samples, via Amazon Mechanical Turk\footnote{\url{https://requester.mturk.com/}.}. Every worker is a native English speaker. 
They are paid 15-20 USD per hour. 
Every worker went through a qualification test of 2 hours and was tested to be highly qualified. We designed one HIT to consist of one question-answer pair.
The average time to complete one HIT (including all steps of the annotation process) is 212.8 seconds. 
After the annotation process, we calculate the inter-annotator agreement rate by the Fleiss’ $\kappa$. Firstly, we computed the Fleiss’ $\kappa$ values across all annotators for the sub-sentence identification task, arriving at a value of 75.97\%. This signifies a robust consensus among the annotators~\cite{moore2006statistics}. Additionally, for the definitive faithfulness score (1-5 Likert Scale), we computed the values involving all annotators and achieved a result of 60.0\%. This concordance among the evaluation participants suggests the human evaluation results are reliable.

We show our human evaluation results and automatic evaluation results in Table~\ref{tab:human_eval}. From this Table, we find that models that perform better in the manual evaluations also have better performance in the automated evaluations. This indicates the high correlation between objective and subjective evaluation.

\begin{table}[]
    \centering
    \begin{tabular}{lcc} \toprule
\bf         Model& \bf  Human & \bf Automatic \\ \midrule
         LLaVA & 0.7708&0.6997\\
         InstructBLIP &  0.7804&0.7165\\
         
         \bottomrule
    \end{tabular}
    \caption{Human evaluation results and automatic evaluation results of different LVLMs on the LLaVA dataset.}
    \label{tab:human_eval}
\end{table}

\begin{table}[t]
    \centering \small
     \resizebox{\columnwidth}{!}{
    \begin{tabular}{lccc}
    \toprule
    \bf Metric & \bf Pearson's $r$ \% &\bf Spearman's $\rho$ \% &\bf Kendall's $\tau$ \% \\ \midrule
        BLEU-1 & -15.1& -10.3 & -7.5 \\
        BLEU-2 &  -12.7& -9.0 & -6.6 \\ 
        BLEU-3 &-7.2 & -10.6 & -7.6 \\ 
        BLEU-4 & -1.9& -8.2 & -5.8 \\ 
        ROUGE-1 & -6.6&-3.0 &  -2.7\\ 
        ROUGE-2 & -5.7&-4.4 &  -3.4\\ 
        ROUGE-L & -8.7&-6.2 &  -4.7\\ 
        METEOR & -12.2&-8.5 & -6.3\\ 
        CHAIR & 16.8&19.2 & 14.8\\ 
        
        CLIP-Score &19.8 & 16.6  & 11.7 \\  
        
        SPICE& 20.2 &21.3 &25.4\\
\midrule
        Ours & 48.17&  38.44& 47.61\\
    \bottomrule
    \end{tabular}
    }
    \caption{Correlation between each evaluation metric and human judgment on LVLM hallucinations, measured by Pearson's $r$, Spearman’s $\rho$, and Kendall’s $\tau$.
    }
    \label{tab:allcorrelation}
\end{table}

\begin{figure}
    \centering
    \includegraphics[width=\linewidth]{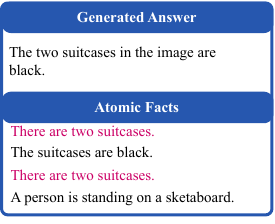}
    \caption{Illustration of atomic facts generated by ChatGPT. The \textcolor{red}{red contents} denote the duplicated atomic fact.} 
    \label{fig:example4facts}
\end{figure}

\begin{figure*}[t]
    \centering
    \includegraphics[width=\linewidth]{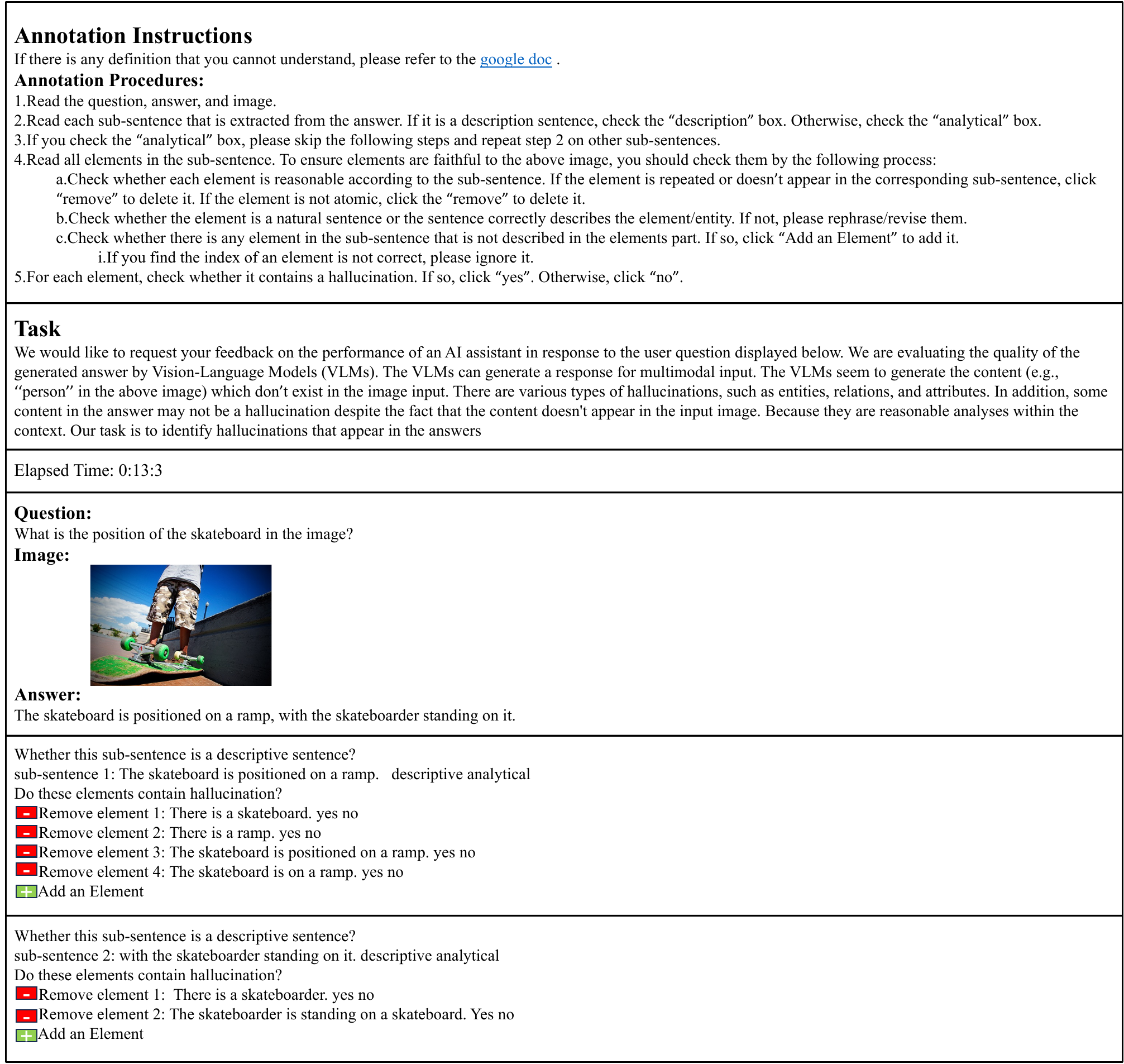}
    \caption{System software User Interface (UI) for annotators. Annotators can read the instructions at the top of the interface and get detailed instructions (see Figure~\ref{fig:instruction}) via a link. Then the annotator can read the task description. In the third part, the annotator can click buttons for sub-sentence identification and atomic fact verification. Meanwhile, they can add, edit, and remove atomic facts to get high-quality atomic information. 
    }
    \label{fig:UI}
\end{figure*}

\begin{figure*}[t]
    \centering
    \includegraphics[width=0.9\textwidth]{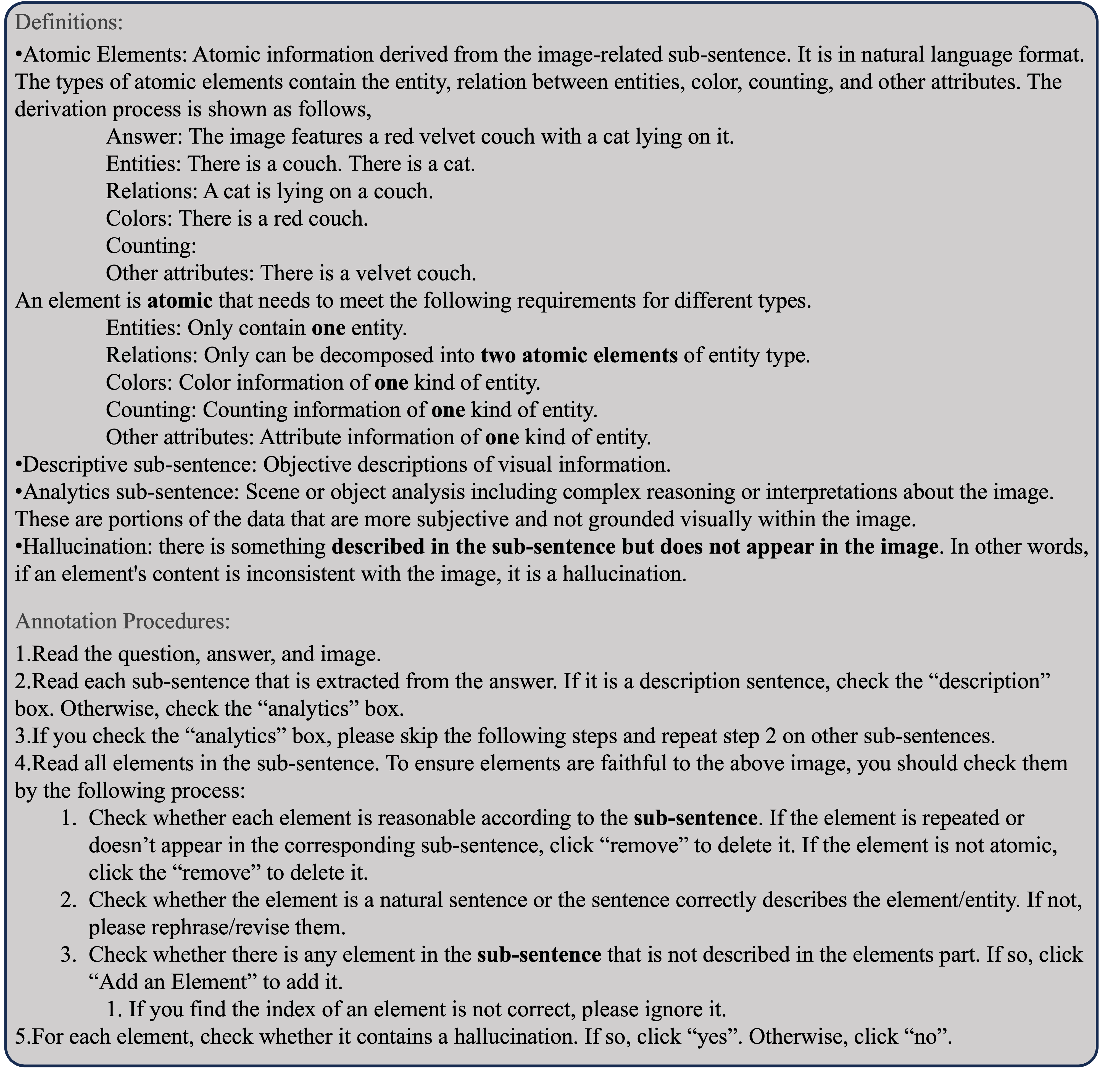}
    \caption{Instructions for data annotation. The instruction includes some definitions (e.g. atomic facts and descriptive sub-sentence) to help annotators understand this task. Meanwhile, it also details the annotation procedures.}
    \label{fig:instruction}
\end{figure*}

To facilitate the annotator's working process, we designed a user interface, as shown in Figure~\ref{fig:UI}. Annotators have the option to start by reading the instructions located at the top of the interface, and they can access more detailed instructions through a link (refer to Figure~\ref{fig:instruction}). Following this, annotators can proceed to review the task description. In the third section, annotators can utilize buttons for sub-sentence identification and atomic fact verification. Simultaneously, they are able to add, modify, or delete atomic facts to enhance the quality of the atomic information. For example, the annotator should remove the duplicated atomic and add entity category fact ``There are suitcases.'' in Figure~\ref{fig:example4facts}.  

Besides, we show a comprehensive correlation comparison in Table~\ref{tab:allcorrelation}. 
Traditional metrics that require references (\ie BLEU, ROUGE, and METEOR), have a poor correlation with human evaluation. For the open-ended question, it is hard to get a ground truth answer. For the reference answer, we use the answers provided by the LLaVA paper. This leads to a poor correlation between these metrics and human evaluation. 

Surprisingly, CLIP-Score shows a similar correlation with CHAIR which is specifically devised for object hallucination evaluation. This demonstrates the robustness and generalization of CLIP-Score.
The original CHAIR show reflects the severity of the hallucinations. The larger the value of CHAIR, the more serious the hallucination problem of the model.
The original CHAIR exhibits a pronounced negative correlation with human evaluation. Hence, we use the negative of CHAIR to compute the correlation. 

Compared with \ours, CHAIR achieves a sub-optimal degree of correlation.  
A potential reason for CHAIR's deviation from human evaluation could be rooted in its inherent design, which narrows its focus predominantly to a limited range of objects. This constrained evaluation scope may not adeptly deal with fine-grained and open-domain hallucinations, thus diminishing its validity and resonance with more comprehensive human evaluations. To justify our viewpoint, we compute the average number of objects with CHAIR for each answer and the result is 2.4, which is far less than the average number of atomic facts (\ie 11.3) found in our human evaluation.  
Amid the varied metrics landscape, our metric \ours\ achieved best correlation with human correlation.

 {We further conduct an ablation study to investigate the overall effect of different VE models and the error introduced by the ChatGPT on \ours. }
Table~\ref{tab:ablation1} reports the correlation between \ours\ calculated by different VE models answers and human answers. We observed that the higher VE model performance is directly related to the human correlation. Table~\ref{tab:ablation2} reports the correlation between \ours\ and different VE models calculated with the annotated atomic facts. Similarly, the higher VE model performance is directly related to the human correlation. 

\section{Error in Atomic Fact Decomposing}
\label{app:decompose}
{To further learn the degree of hallucination in decomposing phrases, we further sample 100 instances from the human evaluation samples and then use ChatGPT to generate atomic facts. Finally, we found the hallucination ratio (hallucinated atomic facts in proportion to all atomic facts) is just 2\%. This verifies the effectiveness of our method.}

\section{Experimental Detail}
We run all VLMs on an NVIDIA A100 GPU. 
{The recognizer accuracy on descriptive
sub-sentence identification task is defined as $a=N_s/N$ where $N$
 is the total number of sentences in the evaluated dataset and $N_s$
 is the number of sentences that are classified correctly.}

\begin{table}[]
    \centering \small
     \resizebox{\columnwidth}{!}{
    \begin{tabular}{lccc}
    \toprule
    \bf Metric & \bf Pearson's $r$ \% &\bf Spearman's $\rho$ \% &\bf Kendall's $\tau$ \% \\ \midrule
        OFA\_EM & 31.85&  21.27& 29.03\\
        BLIP-2-flant5xxl & 41.80& 28.52  & 36.81 \\
        LLaVA-1.5 & 48.17&  38.44& 47.61\\
    \bottomrule
    \end{tabular}
    }
    \caption{Correlation between our ablation methods and human judgment on LVLM hallucinations, measured by Pearson's $r$, Spearman’s $\rho$, and Kendall’s $\tau$.
    }
    \label{tab:ablation1}
\end{table}

\begin{table}[]
    \centering \small
     \resizebox{\columnwidth}{!}{
    \begin{tabular}{lccc}
    \toprule
    \bf Metric & \bf Pearson's $r$ \% &\bf Spearman's $\rho$ \% &\bf Kendall's $\tau$ \% \\ \midrule
        OFA\_EM & 32.34&  22.28& 30.12\\
        BLIP-2-flant5xxl & 45.84& 31.62  & 40.09 \\
        LLaVA-1.5 & 58.46&  42.67& 56.23\\
    \bottomrule
    \end{tabular}
    }
    \caption{Correlation between vem models with the annotated atomic facts and human judgment on LVLM hallucinations, measured by Pearson's $r$, Spearman’s $\rho$, and Kendall’s $\tau$.
    }
    \label{tab:ablation2}
\end{table}

\section{Proportions of the descriptive sub-sentences and analytical sub-sentences }

\begin{figure}
    \centering
    \includegraphics[width=0.9\linewidth] {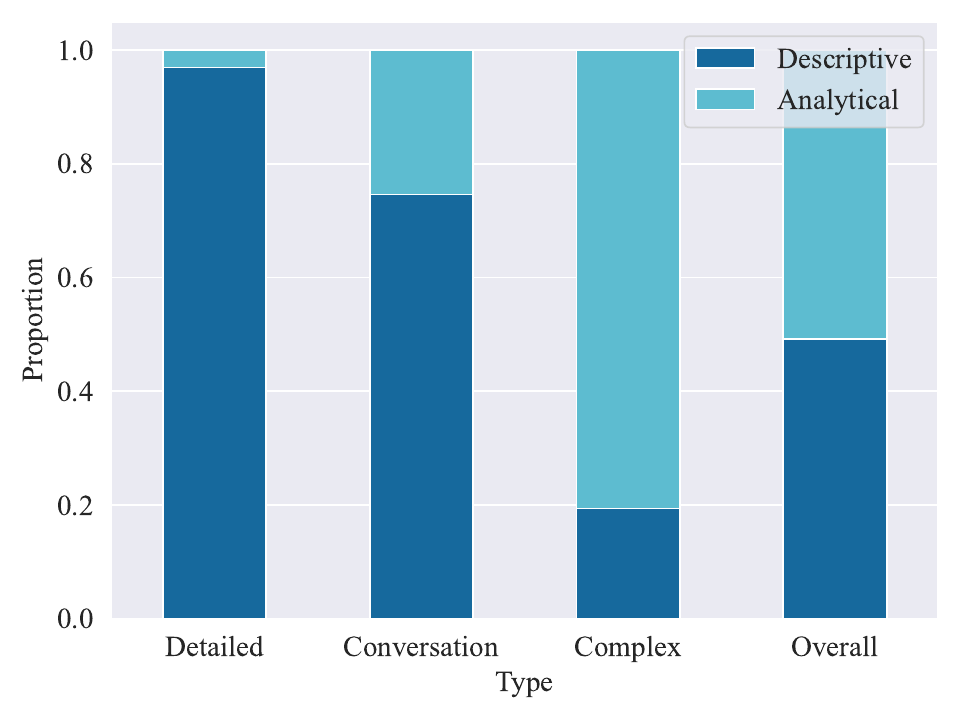}
    \caption{Illustration of the proportions of the descriptive sub-sentences and analytical sub-sentences in the answers. ``Detailed'' and ``Complex'' denote the ``detailed description'' and ``complex question'' categories, respectively. The results are obtained from the 180 annotated samples.}
    \label{fig:proportion}
    \vspace{-2mm}

\end{figure}

To prove the necessity of the sentence identification step, we calculate the proportion of descriptive and analytical sub-sentences in answers to different classes of input questions (Figure~\ref{fig:proportion}). We can observe that the distribution of sub-sentences is significantly different in different category questions. For example, detailed description questions only have a small portion of analytical sub-sentences, while complex questions have the opposite. In addition, analytical sub-sentences account for nearly half of the distribution of clauses in the overall annotated dataset, illustrating the importance of identifying analytical sub-sentences and excluding them from the fact checking step.

\section{Cost and Time Analysis}
\label{app:cost}
{Gathering accurate hallucination evaluation manually for each response is both costly and time-consuming, making it unrealistic.   Therefore, although our reference-free metric typically takes more time than traditional metrics such as BLEU and ROUGE, it is much more important for evaluating model output. 
There is a trade-off between evaluation results and speed. Although some metrics (such as BLEU) can achieve higher speed, our metric can achieve a superior correlation with human evaluation. Compared with human evaluation, our method can speed up the evaluation process. In addition, we can verify multiple facts in parallel to speed up the time. The average consuming time for a long response is 4s, which is much less than the manual time of 212 seconds, reducing about 1.2 USD cost.}

\section{Samples of Description/Analytics Sub-sentence}
We provide three examples of "description"/"analytics" sub-sentence, where [A] denotes the analytical sub-sentence label and [D] denotes the descriptive sub-sentence label.
(1) The skateboard is positioned on a ramp, [D] with the skateboarder standing on it. [D] (2) The image features a white bird, [D] which is likely a swan. [A]
(3) The image features a young boy standing on a skateboard, [D] which is placed on a wooden ramp. [D] The boy is wearing a green shirt and brown shorts, [D] and he is positioned on the ramp, ready to ride down. [D] The wooden ramp is placed on a sidewalk, [D] which is a common location for skateboarding. [A] The presence of the ramp and the sidewalk suggests that the boy is likely practicing or enjoying skateboarding in a public space. [A] The overall composition of the image highlights the boy's focus and determination as he prepares to ride down the ramp, [A] showcasing the excitement and thrill of skateboarding. [A]

\section{Prompts}
\label{prompts}
\begin{figure*}[t]
    \centering
    \includegraphics[width=0.8\textwidth]{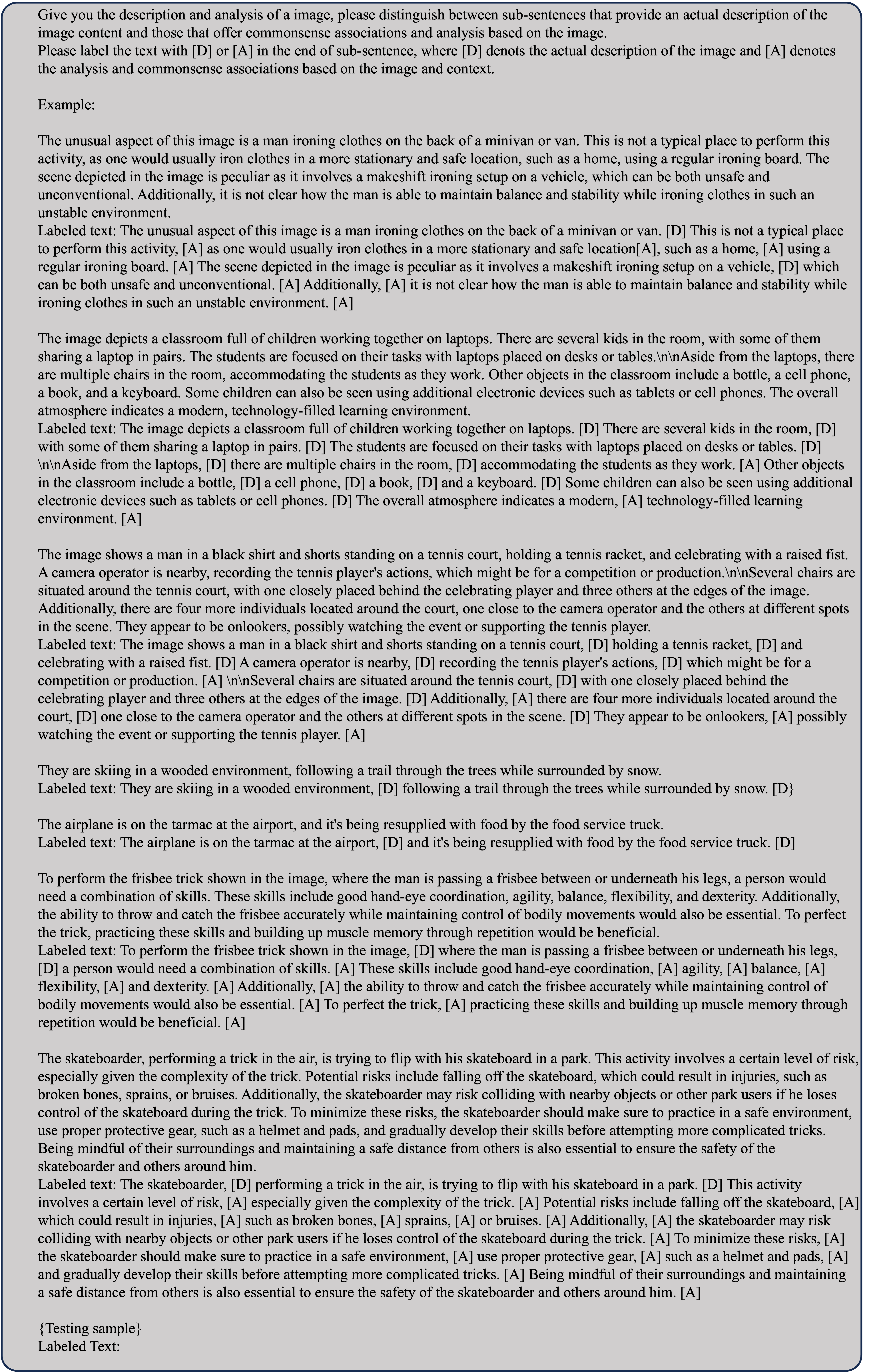}
    \caption{A prompt given to ChatGPT to identify descriptive sub-sentence from answers of VLMs.}
    \label{fig:prompt_des}
\end{figure*}

\begin{figure*}[t]
    \centering
    \includegraphics[width=0.9\textwidth]{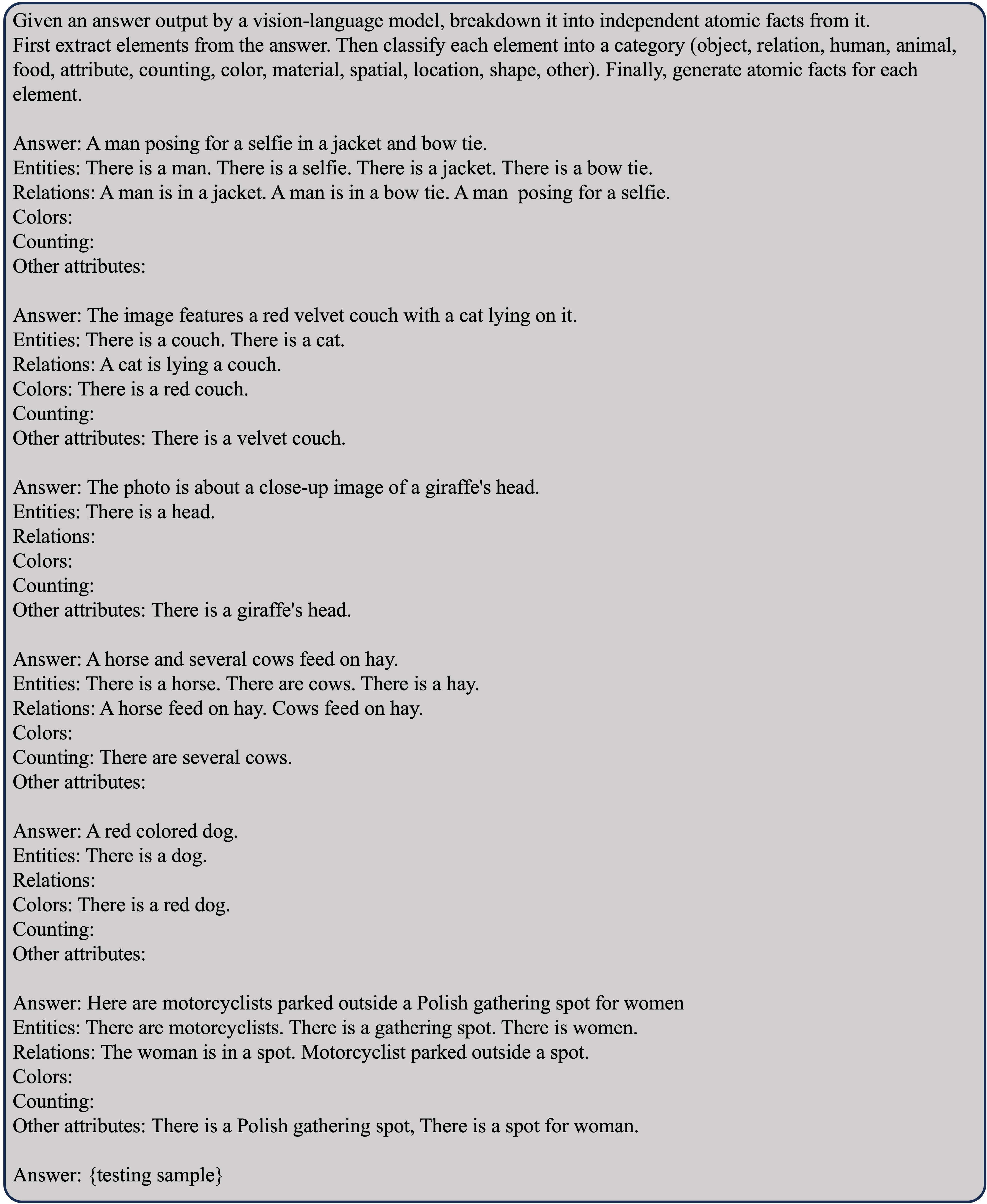}
    \caption{A prompt given to ChatGPT to generate atomic facts of VLMs answers.}
    \label{fig:prompt_decom}
\end{figure*}

We detailed the prompts of sub-sentence identification and atomic fact generation in Figure~\ref{fig:prompt_des} and Figure~\ref{fig:prompt_decom}, respectively.

\section{Results on Closed-source Model}
{To comprehensively evaluate the existing LVLMs, we further evaluate the closed-source model GPT-4V and Gemini~\cite{DBLP:gemini} with several open-source models. The results are shown in Table \ref{tab:gemini}. The results show the closed-source models surpass the existing open-source model. 
}

\begin{table}[]
    \centering \small
     \resizebox{\columnwidth}{!}{
    \begin{tabular}{lccc}
    \toprule
\textbf{Model} & \textbf{LLaVA-1k} & \textbf{Model} & \textbf{LLaVA-1k} \\
        \hline
        Multimodal-GPT & 0.5335 & LLaVA & 0.8360 \\
        MiniGPT-4 & 0.5713 & LLaVA-1.5 & 0.8566 \\
        mPLUG-Owl & 0.7167 & GPT-4V & 0.8788 \\
        InstructBLIP & 0.8091 & Gemini & 0.8865 \\
        \hline
    \end{tabular}
    }
    \caption{Model performance on LLaVA-1k benchmark.    }
\label{tab:gemini}
\end{table}